\documentclass[final]{cvpr}

\makeatletter
\@namedef{ver@everyshi.sty}{}
\makeatother

\usepackage{times}
\usepackage{epsfig}
\usepackage{graphicx}
\usepackage{amsmath}
\usepackage{amssymb}
\usepackage{dsfont}
\usepackage{color}
\usepackage{comment}
\usepackage{subfig}
\usepackage{wrapfig}
\usepackage{nicefrac}
\usepackage{float}
\usepackage[normalem]{ulem}
\usepackage{animate}
\usepackage[export]{adjustbox}
\usepackage{booktabs}
\usepackage{multirow}
\usepackage{enumitem}
\usepackage{tikz}
\usepackage{xspace}
\usepackage{bbold}
\usepackage[font=small]{caption}

\usepackage[pagebackref=true,breaklinks=true,letterpaper=true,colorlinks,bookmarks=false]{hyperref}

\newcommand*\cP{
  \tikz[baseline=(char.base)]{\node[shape=circle,draw=blue,inner sep=1pt](char){\color{blue}$*$};
  }\xspace}
\newcommand*\labelB{
  \tikz[baseline=(char.base)]{\node[minimum width = 10pt, minimum height=12pt,draw](char){B};
  }\xspace}
\newcommand*\labelC{
  \tikz[baseline=(char.base)]{\node[minimum width = 10pt, minimum height=12pt,draw](char){C};
  }\xspace}
\newcommand*\labelQ{
  \tikz[baseline=(char.base)]{\node[minimum width = 10pt, minimum height=12pt,draw](char){Q};
  }\xspace}

\DeclareMathOperator*{\argmax}{arg\,max}

\definecolor{amber}{rgb}{1.0, 0.75, 0.0}

\begin{document}

\title{Binary TTC: A Temporal Geofence for Autonomous Navigation}

\makeatletter
\author{Abhishek Badki$^{1,2}$
  \and
  Orazio Gallo$^1$\\
  $^1$NVIDIA
  \and
  Jan Kautz$^1$\\
  $^2$UC Santa Barbara
  \and
  Pradeep Sen$^2$
}

\maketitle
\begin{abstract}

Time-to-contact (TTC), the time for an object to collide with the observer's plane, is a powerful tool for path planning: it is potentially more informative than the depth, velocity, and acceleration of objects in the scene---even for humans.
TTC presents several advantages, including requiring only a monocular, uncalibrated camera.
However, regressing TTC for each pixel is not straightforward, and most existing methods make over-simplifying assumptions about the scene.
We address this challenge by estimating TTC via a series of simpler, binary classifications.
We predict with \emph{low latency} whether the observer will collide with an obstacle \emph{within a certain time}, which is often more critical than knowing exact, per-pixel TTC.
For such scenarios, our method offers a temporal geofence in $6.4$ ms---over $25\times$ faster than existing methods.
Our approach can also estimate per-pixel TTC with arbitrarily fine quantization (including continuous values), when the computational budget allows for it.
To the best of our knowledge, our method is the first to offer TTC information (binary or coarsely quantized) at sufficiently high frame-rates for practical use.
{\let\thefootnote\relax\footnote{This work was done while A. Badki was interning at NVIDIA.}}
{\let\thefootnote\relax\footnote{Project page: \url{https://github.com/NVlabs/BiTTC}}}
\end{abstract}
\setcounter{footnote}{0}
\vspace{-4mm}
\section{Introduction}\label{sec:intro}

Path planning, whether for robotics or automotive applications, requires accurate perception, which in turn, benefits from depth information.
Many modalities exist to infer depth.
Strategies such as lidar estimate depth accurately but only at sparse locations, in addition to being expensive.
Depth can also be estimated with strategies such as stereo~\cite{scharstein2002taxonomy}, but these introduce issues such as calibration drift over time.

\begin{figure}
    \centering
    \includegraphics[width=0.92\columnwidth]{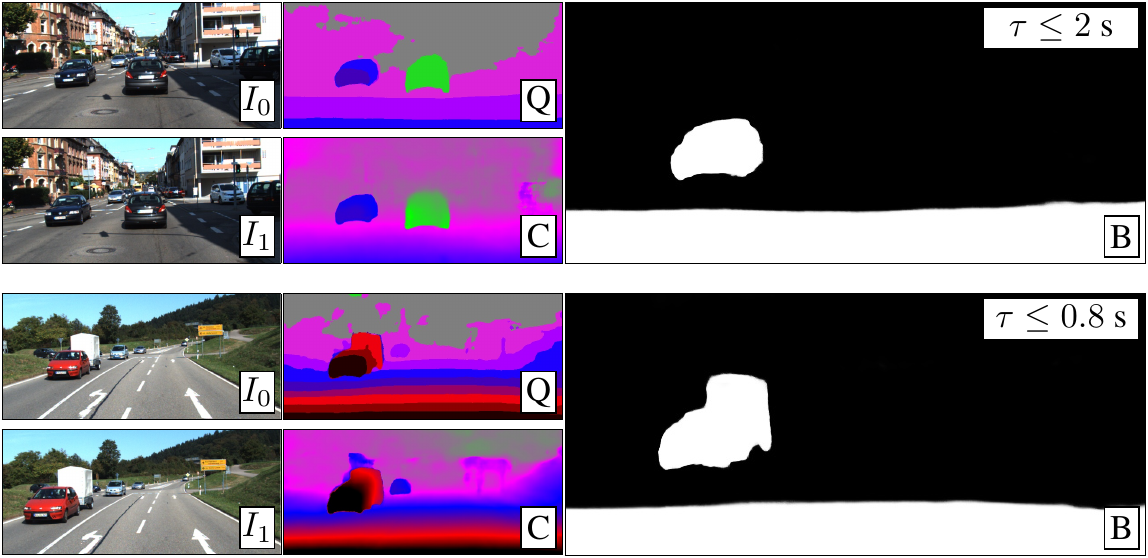}
    \hspace{-1.5mm}
    \raisebox{0.15ex}{\includegraphics[angle=90,origin=l, width=0.069\columnwidth]{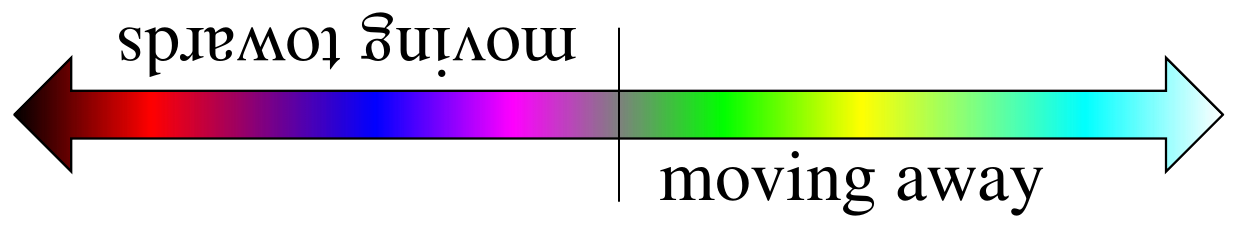}}
    \vspace{-2mm}
    \caption[]{Given $I_0$ and $I_1$, our binary time-to-contact (TTC) estimation acts as a temporal geofence detecting objects that will collide with the camera plane within a given time,\labelB. It only takes $6.4$ ms to compute. Our method can also output quantized TTC,\labelQ, and continuous TTC, \labelC.}\label{fig:teaser}
    \vspace{-5mm}
\end{figure}

An alternative is to use a monocular camera---an attractive, low-cost solution, with light maintenance requirements.
The motion of the camera induces optical flow between consecutive frames, which carries information on the scene's depth.
Depth, however, can only be estimated in the constrained case of static scenes.
For dynamic scenes, the 2D flow of a pixel is a function of its depth, its velocity, and the velocity of the camera.
Disentangling these three components is an under-constrained and challenging problem.
Previous approaches either ignore dynamic regions~\cite{schoenberger2016sfm}, or use strong scene priors~\cite{Luo-VideoDepth-2020,yoon2020novel,zou2018dfnet, yin2018geonet, Luo2020everypixel} to hallucinate their depth.
\emph{Do we really need to disentangle them?}
The role of perception is to inform decisions.
An object moving towards the camera is more critical than another that is potentially closer, but moving away from the camera.
Differently put, predicting the time at which an object would make contact with the camera may be more valuable than knowing its actual depth, velocity, or acceleration~\cite{lee1976ttcForBraking}.

In fact, time-to-contact (TTC), the time for an object to collide with the camera plane under the current velocity conditions, is a traditional concept in psychophysics~\cite{tresilian1991empirical, hecht2004ttc_book} as well as computer vision~\cite{subbarao1990bounds,cipolla1992surface}.
TTC can be estimated from the ratio of an object's depth and its velocity relative to the camera, even when the problem of regressing either one independently is ill-posed.
However, TTC estimation has its own challenges, forcing most of the existing approaches to severely constrain their scope.
For instance, they assume that the scene is static, or that a mask for dynamic objects is provided~\cite{horn2009hierarchical, meyer1994time}.
The recent approach by Yang and Ramanan tackles some of these challenges by learning a mapping between optical flow and TTC directly, thus producing a per-pixel TTC estimate~\cite{yang2020of2ttc}.
However, it relies on accurate optical flow estimation and inherits its limitations, including its heavy computational load.

Unlike most existing approaches, we side-step the need to explicitly compute the optical flow.
Our learning-based approach estimates per-pixel TTC directly from images.
We leverage the relationship between an object's TTC and the ratio of the size of its image in different frames~\cite{cipolla1992surface,horn2007ttc_planar}.
However, because regressing this scale factor exactly is challenging, we focus on whether the size of the object's image is increasing, indicating a collision at some time in the future, or decreasing, indicating that the object is moving away.

More concretely, inspired by Badki~\etal~\cite{badki2020Bi3D}, we perform a series of binary classifications with respect to different scale factors, each corresponding to a specific TTC.
Each classification yields a binary TTC map with respect to the desired time threshold, Figure~\ref{fig:teaser}, insets~\labelB.
Our binary map, efficient to compute, acts as a temporal geofence in front of the camera:
it identifies objects within a given TTC, in $6.4$ ms.\footnote{On an NVIDIA Tesla V100 for 384$\times$1152 images.}
This is useful when a quick reaction time is important.
We can also estimate per-pixel TTC with arbitrary quantization, as shown in Figure~\ref{fig:teaser}, insets~\labelQ, or continuous, Figure~\ref{fig:teaser}, insets~\labelC.
These different levels of quantization, from binary to continuous, can be predicted with the same core network.
In fact, quantization levels can be added, removed, or moved dynamically at inference time, based on the current needs of the autonomous agent.
Given the scarcity of TTC ground truth data, to impose additional inductive bias to our network we also introduce binary optical flow estimation as an auxiliary task.
We achieve competitive performance for TTC estimation against existing methods, even stereo-based methods, but we are from $25\times$ to several orders of magnitude faster.

\section{Related Work}\label{sec:related}

While valuable for navigation, 3D information about the scene is challenging to gather with a monocular camera.
Existing methods assume the scene to be static~\cite{engel2014lsd, ummenhofer2017demon, huang2018deepmvs, liu2019neural}, or estimate single-image \emph{relative} depth, rather than \emph{metric} depth~\cite{eigen2014depth, garg2016unsupervised, godard2017unsupervised, fu2018deep, ranftl2020towards}.
Single-image relative depth can be ``upgraded'' to metric by computing the optical flow between multiple images~\cite{Luo-VideoDepth-2020,yoon2020novel,zou2018dfnet, yin2018geonet, Luo2020everypixel}, but this requires strong priors, and it is brittle for complex, large motions.
Nevertheless, with depth maps and optical flow we can estimate scene flow, which captures both depth and velocity~\cite{schuster2018dense}.
A recent approach by Hur and Roth elegantly combines these principles by learning scene priors to decompose the depth and the velocity directly from the estimated optical flow information~\cite{hur2020selfmonosf}.

Instead of regressing depth and velocity, we focus on estimating their ratio directly from images, which yields time-to-contact (TTC).
We do this via multiple binary classifications.
Here we discuss the state-of-the-art in terms of these two axes---TTC and regression via classification.

\subsection{Time-to-Contact}
Time-to-contact (TTC) was studied in psychophysics and psychology, even before it attracted the attention of the computer vision community.
Early work by Lee, for instance, suggested that TTC is sufficient for making decisions about braking, and is likely to be picked up by the driver faster than distance, speed, or acceleration of objects in the scene~\cite{lee1976ttcForBraking}.
From a computational standpoint, TTC is appealing because it only depends on the ratio of depth and velocity, which can be estimated directly from images, even when estimating either one is an ill-posed problem~\cite{horn2009hierarchical}.
Several traditional approaches have been proposed to estimate TTC from the estimated optical flow~\cite{meyer1992estimation, meyer1994time, camus1995calculating}.
Horn~\etal proposed a direct method for TTC estimation that only uses the constant brightness assumptions~\cite{horn2007ttc_planar,horn2009hierarchical}.
While these approaches estimate the TTC for an object that is moving relative to the camera, they require masks for dynamic objects, which limits their practical impact.

We propose a learning-based approach for TTC estimation that handles multiple dynamic objects---without needing any segmentation---and estimates per-pixel TTC.
The elegant, closely related work by Yang and Ramanan estimates optical flow, uses it to compute the scaling factor of objects, and maps it to TTC~\cite{yang2020of2ttc}.
Xu \etal also estimate the scaling of objects, but do so by modeling the change of objects' size explicitly for optical flow estimation~\cite{xu2012scaleinvof}.
However, computing the full optical flow is time-consuming, and estimating TTC from optical flow inherits its limitations.
Instead, following Horn~\etal~\cite{horn2009hierarchical}, we compute TTC directly from the input images, side-stepping optical flow computation altogether.
Moreover, inspired by Badki~\etal~\cite{badki2020Bi3D}, we solve TTC estimation via a series of binary classifications.
Each binary classification can be computed independently of the others at over $150$ fps.
This can be thought of as a temporal geofence, detecting pixels or objects within a given TTC.
For existing methods, including the method by Yang and Ramanan~\cite{yang2020of2ttc}, this can only be achieved by computing the TTC for all the pixels or objects in the scene, and then thresholding it.

\begin{figure*}
    \vspace{-5mm}
    \centering
    \subfloat[Inputs]       {\includegraphics[height=80pt, trim={0.04in 0.05in 2.85in 0}, clip]{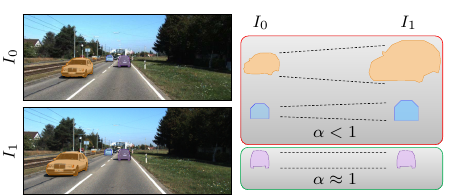}}
    \subfloat[Scaled inputs]{\includegraphics[height=80pt, trim={1.9in 0.05in 0.9in 0}, clip]{figures/intuition/intuition_new_with_bg.pdf}}
    \subfloat[Temporal geofence]{\includegraphics[height=80pt, trim={3.84in 0.05in 0.05in 0}, clip]{figures/intuition/intuition_new_with_bg.pdf}}
    \vspace{-2mm}
    \caption{\textbf{Intuition.}
    Given two images of a dynamic scene, $I_0$ and $I_1$, we define a temporal geofence to detect objects expected to cross the camera plane before a given time $\tau_i$ from the time of capture of $I_0$.
    Compare $I_0$ and $I_1$, (a).
    The images of the orange and blue cars are smaller in $I_0$ ($\alpha<1$), while the image of the purple car is roughly unchanged.
    This allows us to predict that only the first two cars will collide with the camera plane.
    We propose to perform this comparison after scaling the source image by a factor $\alpha_i$ (corresponding to TTC value $\tau_i$).
    The orange car is still larger in $I_1^{\alpha_i}$ ($\alpha<\alpha_i$), indicating that it will cross the camera plane \emph{before} the specified $\tau_i$.
    On the other hand, the blue and purple cars will not.
    Rather than regressing the exact scale factor, we classify a pixel as making contact before or after $\tau_i$ by classifying if the objects scale up or down in $I_1^{\alpha_i}$ with respect to $I_0$. 
    This yields a binary TTC probability map for $\tau_i$, (c).}
    \label{fig:intuition}
    \vspace{-3mm}
\end{figure*}

\subsection{3D Inference as a Classification Problem}
The idea of posing 3D regression as a classification task has a rich history.
Several learning-based approaches estimate depth via a multi-class classification task~\cite{kendall2017end, chang2018pyramid, zhang2019ga, im2019dpsnet}.
These are more accurate than other learning-based approaches that pose the problem as a regression task~\cite{Menze2015kittidataset,dosovitskiy2015flownet}.
Badki~\etal introduced a method that allows us to control the trade-off between latency and accuracy~\cite{badki2020Bi3D}.
Instead of posing depth as a multi-class classification problem, they solve it via multiple binary classifications.
Each classification provides useful information about the scene at high frame-rates.
Our approach is based on the same intuition.
We are the first learning-based approach to estimate per-pixel TTC directly from the input images, and to pose it as a (binary) classification problem.

\section{Method}\label{sec:overview}

Despite some time-to-contact (TTC) estimation methods dating back to the 1990s~\cite{subbarao1990bounds,cipolla1992surface}, TTC never rose to the popularity of other techniques that are now mainstream, such as optical flow or stereo estimation.
This is due in part to its intrinsic limitations and to the challenges it poses, which we discuss below.
Note that in the rest of the paper we often attribute the properties of objects to the corresponding pixels, to simplify the discussion.
For instance, we talk of ``a pixel's velocity'' to indicate the projection on the image plane of the velocity of the object imaged by that pixel.

\subsection{A Review on Time-to-Contact}\label{sec:ttc_overview}

Consider two frames of a static scene captured by a moving camera.
Pixel-level correspondences allow us to compute depth, if the camera information is known.
In the more realistic case of dynamic scenes, however, the problem becomes ill-posed:
the displacement of a pixel is the result of its depth, its velocity, and the camera velocity, all of which cannot be disambiguated without strong priors.
In this case, the concept of time-to-contact (TTC) comes to the rescue.
Given an object $\mathcal{O}$, the TTC $\tau$, \ie, the time at which object $\mathcal{O}$ will (or did) cross the camera plane, can be written as
\begin{equation}\label{eq:ttc_from_depth_ratio}
    \tau = - Z_{\mathcal{O}} \bigg/ {\frac{dZ_{\mathcal{O}}}{dt}} = - Z_{\mathcal{O}} / \dot{Z}_{\mathcal{O}},
\end{equation}
where the origin is at the camera, and we assume that the current velocity conditions will continue.
$Z_{\mathcal{O}}$ is the depth of the object from the camera plane, and $\dot{Z}_{\mathcal{O}}$ its relative velocity.
Equation~\ref{eq:ttc_from_depth_ratio} shows the first appealing feature of TTC:
even if the depth and the velocity of the object cannot be estimated independently, $\tau$ can be computed from their ratio.

However, we are interested in computing the TTC from pixel displacements alone.
To do that, we need an additional piece of information: the location of the focus-of-expansion (FOE).
Given two frames of a static scene captured by translating the camera, all the pixels in the image move along lines originating from the FOE, the image of the point towards which the camera is moving.
Under the same assumptions, the FOE coincides with the epipole.
The relationship between the TTC, $\tau$, and the velocity of the pixel is given by
\begin{equation}\label{eq:ttc_from_velocity}
    \dot{x} = \frac{x-x_0}{\tau} \quad \text{and} \quad \dot{y} = \frac{y-y_0}{\tau},
\end{equation}
where $(x_0,y_0)$ is the FOE.
Equation~\ref{eq:ttc_from_velocity} can be easily derived by differentiating the projection of a 3D point onto the image plane with respect to time~\cite{horn2007ttc_planar}.
Note that the velocity $(\dot{x}, \dot{y})$ can be computed from the optical flow $(u,v)$, which allows us to write
\begin{equation}\label{eq:ttc_from_flow}
    \tau = \frac{x-x_0}{u} \cdot T \quad \text{and} \quad \tau = \frac{y-y_0}{v} \cdot T,
\end{equation}
where $T$ is the time elapsed between the two frames.

Equation~\ref{eq:ttc_from_flow} shows a second compelling reason to use TTC:
there is no need for camera calibration.\footnote{We assume a pin-hole camera model.}
There are also challenges, however.
If a rigid object is dynamic and translates with respect to the scene, its pixels move along lines centered around a different FOE. To estimate the TTC using Equation~\ref{eq:ttc_from_flow}, then, we need to localize the FOE of \emph{each} dynamic object.
Moreover, if an object is deformable, the FOE varies with each pixel, and for general motion, we need to further compensate for rotations.
This is why traditional approaches had to rely on oversimplifying assumptions.

However, we can compute the size of the image of a fronto-parallel, planar, non-deformable object of size $S_{\mathcal{O}}$ at distance $Z_{\mathcal{O}}$ as~\cite{horn2007ttc_planar}
\begin{equation}\label{eq:scale_Z_inverse_relation}
    s_{\mathcal{O}} = fS_{\mathcal{O}}/Z_{\mathcal{O}},
\end{equation}
where $f$ is the focal length of the camera.
Since $f$ and $S_{\mathcal{O}}$ are constant, differentiating Equation~\ref{eq:scale_Z_inverse_relation} yields
\begin{equation}\label{eq:Z_to_scale}
    Z_{\mathcal{O}}/\dot{Z}_{\mathcal{O}}= -s_{\mathcal{O}}/\dot{s}_{\mathcal{O}},
\end{equation}
which, plugged into Equation~\ref{eq:ttc_from_depth_ratio}, allows us to estimate the TTC using only information about the size of an object's image and its rate of change:
\begin{equation}\label{eq:ttc_from_scale}
    \tau = s_{\mathcal{O}}/\dot{s}_{\mathcal{O}}.
\end{equation}
Note that both Equation~\ref{eq:ttc_from_flow}~and~\ref{eq:ttc_from_scale} effectively look at scaling.
However, the latter is independent of the point with respect to which the scaling is performed, whereas for the former, that point is the FOE.
Of course, under more realistic conditions (\eg, not planar or not fronto-parallel objects), Equation~\ref{eq:ttc_from_scale} becomes an approximation, which requires proper handling, as we show later.

\begin{figure}
    \vspace{-4mm}
    \centering
    \input{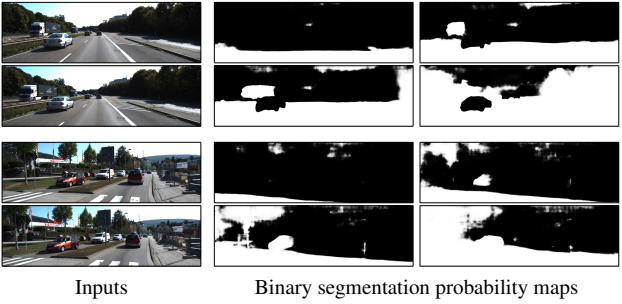}
    \caption{\textbf{Binary TTC maps.} Our method can directly identify the pixels that will contact the camera plane before a given time. From top to bottom, we show results for four TTC values for a case of highway driving and for a camera that rotates.}\label{fig:binary_results}
    \vspace{-4mm}
\end{figure}

\subsection{TTC via Multiple Binary Classifications}\label{sec:ttc_as_classification}
In this section, we first provide the intuition motivating our method.
We then describe binary TTC, the core of our method, which detects pixels predicted to collide with the camera plane within a given time.
We further show how an arbitrary number of binary classifications can be combined to estimate, for each pixel, a quantized version of the TTC.
To simplify the description, here we assume positive TTCs, \ie, we focus on objects moving towards the camera rather than away from it.
In Section~\ref{sec:implementation}, we describe a small adaptation of our method that allows us to seamlessly remove this distinction.

\subsubsection{Intuition}\label{sec:ttc_intuition}
Given two images captured at times $t_0$ and $t_1$, Equation~\ref{eq:ttc_from_depth_ratio} can be approximated as
\begin{equation}\label{eq:ttc_from_depth_discrete}
    \tau = - Z(t_0) \bigg/ {\bigg(\frac{Z(t_1) - Z(t_0)}{t_1 - t_0}\bigg)} = \frac{t_1 - t_0}{1-\frac{Z(t_1)}{Z(t_0)}}.
\end{equation}
If we assume fronto-parallel and planar objects that do not rotate, we can combine Equations~\ref{eq:ttc_from_depth_discrete}~and~\ref{eq:scale_Z_inverse_relation}, thus expressing the TTC as a function of observations in image space:
\begin{equation}\label{eq:ttc_from_scale_discrete}
    \tau = \frac{t_1 - t_0}{1-\frac{s(t_0)}{s(t_1)}} = \frac{t_1 - t_0}{1-\alpha},
\end{equation}
where $s$ is the size of the image of an object or region in the scene.
In other words, with Equation~\ref{eq:ttc_from_scale_discrete}, the TTC can be computed from the scale factor $\alpha$.
This simplifies our task, compared with having to estimate per-pixel FOEs and optical flow.
However, regressing $\alpha$ for each pixel explicitly, which requires defining the size of objects or image regions and tracking its change over time, is not straightforward.
Instead, we consider a pair of images, $I_0$ and $I_1$, and compute $I^{\alpha_i}_1$, a version of $I_1$ scaled by a factor $\alpha_i$.
This scaling factor $\alpha_i$ corresponds to a unique TTC $\tau_i$.
The regions whose size matches between $I_0$ and $I^{\alpha_i}_1$ will collide with the camera plane exactly at $\tau_i$.
Furthermore, we can expect regions that are larger (or smaller) in $I^{\alpha_i}_1$ to collide with the camera before (or after) $\tau_i$, see Figure~\ref{fig:intuition}.

Instead of regressing the scale factor $\alpha$ directly, then, we propose to train a neural network to take such pairs of images and classify regions in $I^{\alpha_i}_1$ as being larger or smaller than the corresponding regions in $I_0$, where the correspondence is learned implicitly.
Our approach is inspired by the work of Badki~\etal~\cite{badki2020Bi3D} for stereo.
They also learn to classify the disparity of a pixel as being larger or smaller than a given disparity, instead of regressing the disparity directly.

\begin{figure}
    \vspace{-1mm}
    \centering
    \includegraphics[width=\columnwidth]{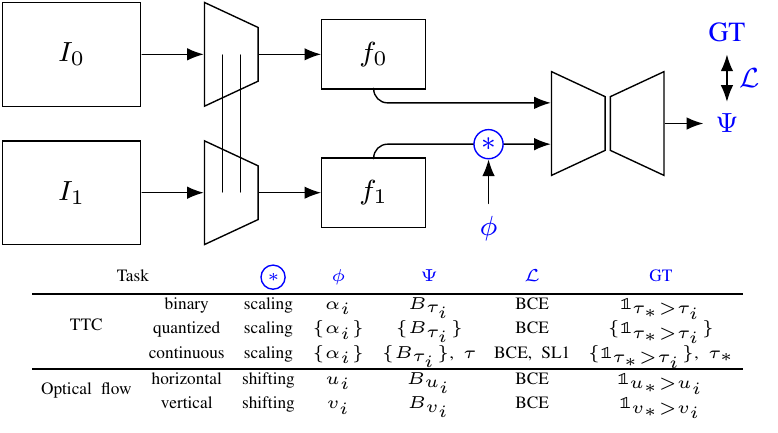}
    \vspace{-6mm}
    \caption[]{\textbf{Architecture.} We perform all of our tasks with minor modifications to the same backbone. We preprocess the input images by extracting features and applying a task dependent operation, \cP, to the features of the second image. The input parameter {\color{blue}$\phi$}, the loss, {\color{blue}$\mathcal{L}$}, the ground truth, {\color{blue}GT}, and the output, {\color{blue}$\Psi$} for each task are listed in the table. Note that $\{\cdot\}$ indicates a set, and $\mathbb{1}$ the indicator function.}
    \label{fig:arch}
    \vspace{-4mm}
\end{figure}

\subsubsection{Binary TTC}\label{sec:ttc_binary}
The inputs to our method are two images and scale factor $\alpha_i$ corresponding to the TTC we want to analyze.
We first extract features $f_0$ and $f_1$ from both images, and scale $f_1$ by $\alpha_i$, to obtain $f_1^{\alpha_i}$.
Rather than classifying the objects as scaling up or down between $f_0$ and $f_1^{\alpha_i}$, we train a lightweight network to directly classify whether each pixel's TTC is larger or smaller than $\tau_i$, using a binary cross-entropy loss.
That is, the network predicts a probability map
\begin{equation}\label{eq:binary_map}
    B_{\tau_i}(x,y) = p(\tau(x,y)>\tau_i; f_0, f_1^{\alpha_i}),
\end{equation}
which can be binarized by simple thresholding.
We train directly to predict binary TTC instead of explicitly detecting if the size of the image of objects is getting larger or smaller with respect to $\alpha_i$ for two reasons.
First, ground truth data for the scale factor $\alpha$ is challenging to even define beyond a small neighborhood of pixels, unless objects move rigidly and only translate.
Moreover, as discussed before, Equation~\ref{eq:ttc_from_scale_discrete} is an approximation in the common case of objects that are not planar, and for non-rectilinear motions.
Equation~\ref{eq:ttc_from_depth_discrete}, which establishes the relationship between the change in depth and the TTC, allows us to generate the ground truth data from existing datasets, as we explain in Section~\ref{sec:data_ttc_from_sceneflow}.
A network trained to classify the TTC directly can learn the necessary priors to compensate for the approximations introduced by Equation~\ref{eq:ttc_from_scale_discrete}.
Our core architecture is shown in Figure~\ref{fig:arch}.
This very architecture is also used for all the tasks we describe below, with the caveat that the terms in {\color{blue}blue} differ for each task.

\subsubsection{Quantized TTC}\label{sec:ttc_quantized}
Our core approach naturally extends to estimating a coarsely quantized TTC for each pixel, which may be more useful than binary in certain scenarios.
We note that Equation~\ref{eq:binary_map} is a complementary cumulative distribution function.
Therefore, for two time-to-contact values, $\tau_j > \tau_i$, we can compute
\begin{equation}\label{eq:p_quantized}
    p(\tau_i < \tau(x,y) \le \tau_j) = B_{\tau_i}(x,y) - B_{\tau_j}(x,y).
\end{equation}
Consider a set of TTC values $\{\tau_i\}_{i=1:N}$, and assume they are in increasing order.
After computing Equation~\ref{eq:binary_map} for each of the $N$ TTC values, we can estimate the quantization bin in which the TTC of a pixel falls as
\begin{equation}\label{eq:quantized_map}
    Q(x,y) = \argmax_i \bigg( p\big(\tau_i < \tau(x,y) \le \tau_{i+1}\big) \bigg).
\end{equation}
The different TTC values can be spaced non-uniformly.

\subsubsection{Continuous and Selective TTC}\label{sec:ttc_continuous}
While our method is specifically designed for binary and quantized TTC estimation, it can also estimate per-pixel, continuous TTC.
In principle, we can approximate continuous values by predicting quantized TCC (Section~\ref{sec:ttc_quantized}) with a larger set of TTC values $\{\tau_i\}_{i=1:N}$.
This, however, fails to exploit the relationship between the binary classifications for different $\tau_i$'s, for a pixel.
We slightly modify the approach while still preserving its binary classification core, as shown in Figure~\ref{fig:arch}.
Specifically, instead of taking consecutive pairs of probability maps and applying Equation~\ref{eq:p_quantized}, we stack them.
\begin{wrapfigure}[4]{r}[0pt]{0.38\columnwidth}
    \hspace{-6mm}
    \includegraphics[width=0.44\columnwidth]{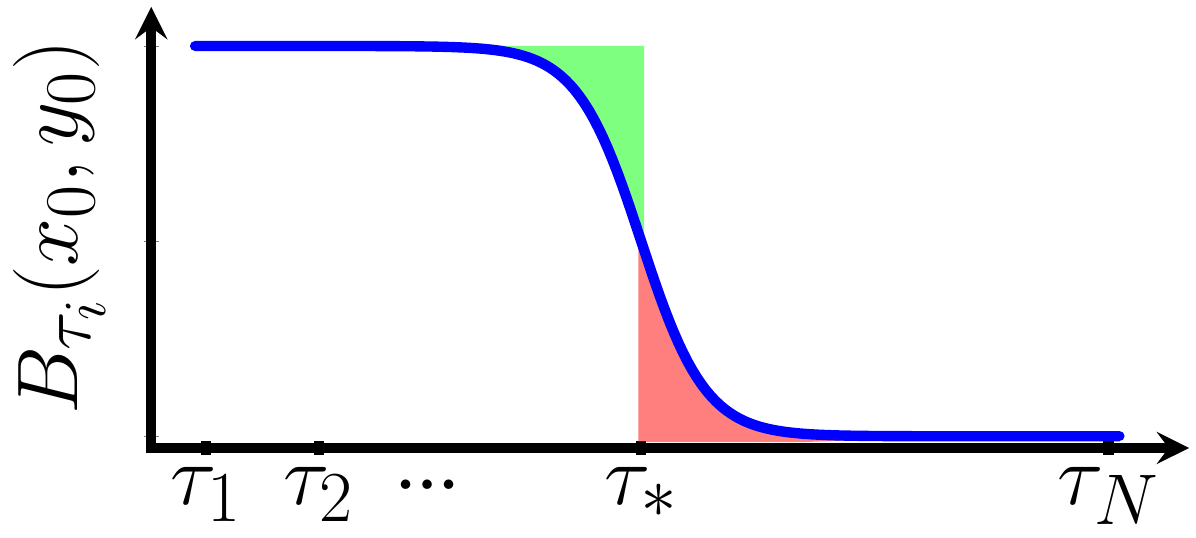}
\end{wrapfigure}
For a specific pixel $(x_0,y_0)$ we generally see a progression as in the plot on the right.
That is, $B_{\tau_i}(x_0,y_0)$ (the probability that the object will collide \emph{after} $\tau_i$) is consistently high for $\tau_i \ll \tau_*$, and low for $\tau_i \gg \tau_*$, where $\tau_*$ is the correct TTC value.
In the transition region, the network is uncertain, which is why aggregating information across multiple $\tau_i$ is beneficial.
Badki~\etal~\cite{badki2020Bi3D}, who obtain a similar curve for disparity, propose to estimate the transition point, $\tau_*$ in our case, by computing the area under the curve (AUC),
\begin{equation}\label{eq:AUC}
    \text{AUC}(x,y) = \sum_i (\tau_{i+1} - \tau_i) \cdot B_{\tau_i}(x,y).
\end{equation}
To understand why Equation~\ref{eq:AUC} yields the desired result, consider the case of a step function, \ie, $\tau_*$ aligns with a quantization boundary: AUC$=\tau_*\cdot 1=\tau_*$.
This relationship holds for the more common case of a smooth transition region:
because the transition is generally symmetric around $\tau_*$, the red and green areas in the plot have similar extent and compensate for each other.
Because the AUC is differentiable, we can use it to fine-tune our network so that combining a set of probability values using the AUC operation yields continuous TTC values.

\subsubsection{A Note on Inference-Time Tradeoff}\label{sec:ttc_discussion}
Binary, quantized, and continuous TTC estimation yield increasingly rich information for navigating an environment.
We note that, when estimating quantized and continuous TTC, the multiple binary classifications can be run in parallel, as they are independent of each other.
Therefore we can compute quantized and continuous TTC at roughly the same frame-rate as for binary quantization---just above $150$ fps.
If multiple binary classifications cannot be run in parallel due to hardware limitations, the cost for computing quantized and continuous TTC grows linearly with the number of levels.
In such cases, one can decide the number of quantization levels dynamically to best leverage the trade-off between accuracy and latency.
For situations where a fast response time is critical, such as highway driving, binary TTC may be sufficient.
For slower navigation (\eg, for robotics or for parking lot driving), our method can be adapted to trade latency for a finer quantization \emph{at inference time}.

\subsection{Dealing with the Lack of Training Data}\label{sec:training_with_of}
\begin{figure}
    \vspace{-3mm}
    \centering
    \input{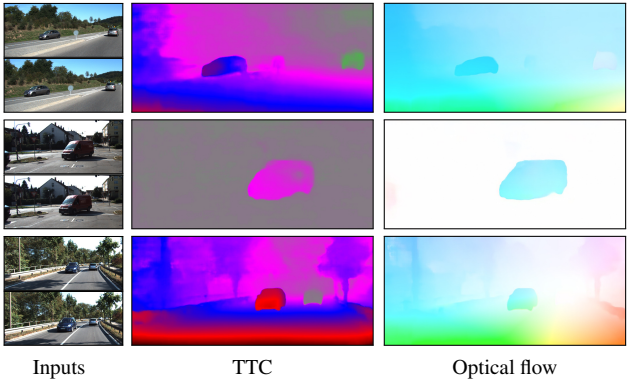}
    \vspace{-2mm}
    \caption{\textbf{Optical flow as an auxiliary task.} We estimate optical flow as an auxiliary task to improve our TTC estimation. Like for continuous TTC, we estimate optical flow via a series of binary classifications. While optical flow is not a goal for us, this figure shows that our prediction is reasonable.}\label{fig:optical_flow}
    \vspace{-4mm}
\end{figure}

As for any learning-based method, having access to a large amount of data is critical to properly train our network.
Unfortunately, there are no datasets with TTC ground truth data and few scene flow datasets that we can use to infer it (Section~\ref{sec:data_ttc_from_sceneflow}).
To increase the training data, we leverage a closely related task: binary optical flow estimation.
We use the same binary approach, but we shift the features (horizontally and vertically) rather than scaling them.
We then classify the direction of the shift (left/right or up/down).
For horizontal shifts, we seek to predict a probability map relative to a given shift $u_i$
\begin{equation}\label{eq:of_u_binary}
    B_{u_i}(x,y) = p(u(x,y)>u_i; f_0, f_1^{u_i}),
\end{equation}
where $f_1^{u_i}$ are the features extracted from the second image and shifted by $u_i$.
The equation for the vertical shift $v_i$ is analogous.
Using optical flow to pre-train our network and continuing using it as an auxiliary task yields a stronger inductive bias, as we discuss in Section~\ref{sec:results}.
Although optical flow is an auxiliary task, we show results in Figure~\ref{fig:optical_flow} to allow for a visual evaluation.

\section{Implementation Details}\label{sec:implementation}
\subsection{TTC Data Generation}\label{sec:data_ttc_from_sceneflow}
A dataset providing per-pixel TTC ground truth does not exist.
However, we can use ratio of the depth in different frames to compute the TTC with Equation~\ref{eq:ttc_from_depth_discrete}.
This ratio can be computed from datasets offering per-pixel scene flow $[X(t_0),Y(t_0),Z(t_0)] \rightarrow [X(t_1),Y(t_1),Z(t_1)]$.
To train for TTC estimation we use the Driving dataset from the SceneFlowDatasets~\cite{mayer2016large}.
We also use the KITTI15~\cite{Menze2015kittidataset} dataset to train for TTC estimation.
We split the training dataset of KITTI15 into train and validation split, and show analysis on the validation part of the dataset.
To pre-train our network for the task of binary optical flow, we use the FlyingChairs2~\cite{dosovitskiy2015flownet, iig2018generic} and the FlyingThings3D~\cite{mayer2016large} datasets.
We also use optical flow data available from Driving~\cite{mayer2016large} and KITTI15~\cite{Menze2015kittidataset}, and train our network for both binary optical flow and TTC estimation.

\subsection{Working in the Inverse TTC Domain}\label{sec:inverse_ttc_domain}
We are interested in objects predicted to collide within $\tau_i$ seconds.
However, $\tau \in (-\infty,\infty)$, with positive and negative values indicating objects moving towards and away from the camera, respectively.
That is, some of the pixels whose TTC is smaller than $\tau_i$ will never collide with the camera plane.
In practice, then, we seek to identify the pixels such that
\begin{equation}\label{eq:interesting_ttc_range}
    (\tau(x,y) \leq \tau_i) ~~ \cap ~~ (\tau(x,y)>0),
\end{equation}
which introduces an additional complication.
Moreover, the effect of TTC in the image space in not linear.
A simple solution is to work with the ratio-of-depths, the inverse of the TTC domain:
\begin{equation}\label{eq:ratio_of_depths}
    \eta = \frac{Z(t_1)}{Z(t_0)} = 1 - \frac{t_1 - t_0}{\tau}.
\end{equation}
Yang and Ramanan termed it motion-in-depth~\cite{yang2020of2ttc}.
Thanks to Equation~\ref{eq:ratio_of_depths}, the effect of TTC is linear in image space and Equation~\ref{eq:interesting_ttc_range} simply reduces to
\begin{equation}\label{eq:eta}
    \eta(x,y)\leq \eta_i.
\end{equation}
For binary TTC, then, we simply scale the features of the source image by a factor $\alpha_i = \eta_i$.

Similarly, for continuous estimation, we sample planes uniformly in the inverse TTC domain.
We apply uniformly spaced scale factors to the features of the source image: $\{\alpha_i\}_{i=1:N} = \{\alpha_0 + i\Delta\alpha\}_{i=1:N}$.
The AUC of the resulting segmentation gives us the map $\eta(x,y)$, which we then covert to a TTC map.

\begin{figure*}
    \vspace{-3mm}
    \centering
    \input{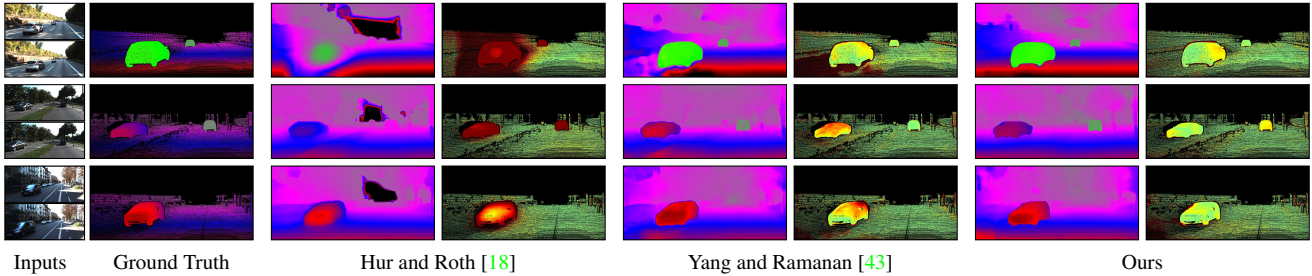}
    \vspace{-2mm}
    \caption{\textbf {Comparison on continuous TTC estimation.} We show the predicted TTC and the error map (red and green indicate high and low error, respectively) for all methods. Our method and Yang and Ramanan's method explicitly train for TTC estimation and provide better results than the monocular scene flow method of Hur and Roth. Note the lower error in the TTC estimation for our method.}\label{fig:all_results}
    \vspace{-3mm}
\end{figure*}

\subsection{Architecture}\label{sec:architecture}
Figure~\ref{fig:arch} shows our architecture.
The feature extraction module uses spatial pyramid pooling layers as PSMNet~\cite{chang2018pyramid}.
The resulting 32-channel feature maps are at one-third the input image resolution.
We then apply a task-dependent operator,\cP, parametrized by {\color{blue}$\phi$}, to $f_1$ and generate $f_1^\phi$.
For instance, when estimating the horizontal component of binary optical flow,\cP shifts the features by a horizontal shift $u_i$, and for binary TTC, it scales them by $\alpha_i$.
The full list for each task is in the table in Figure~\ref{fig:arch}.
To avoid cropping out features, we zero-pad $f_0$ and $f_1^\phi$ to $1.5\times$ the feature map resolution.
The concatenation of $f_0$ and $f_1^\phi$ is fed to a 2D encoder-decoder network with skip connections.
This module has three heads---one for binary TTC, one for binary horizontal optical flow, and one binary vertical optical flow.
We apply a sigmoid to each output to obtain the respective probability maps.
Intuitively, this network tells us if the features in $f_1^\phi$ are scaling up/down, shifting right/left, or shifting up/down with respect to the corresponding features in $f_0$.

\subsection{Training}\label{sec:training}
We pre-train our network for the binary optical flow task on the FlyingChairs2 dataset and train it further on the FlyingThings3D dataset~\cite{mayer2016large} using a binary cross-entropy (BCE) loss with respect to a thresholded version of the ground truth.
The TTC head is left unsupervised in this stage.
We then fine-tune our network for estimating binary TTC first on the Driving~\cite{mayer2016large} and then on the KITTI15~\cite{Menze2015kittidataset} datasets.
Since both datasets also offer optical flow data, in this second stage we  train for both binary optical flow and binary TCC.
In Section~\ref{sec:results} we discuss the impact of this choice on the quality of the results.
We use relative weights of $0.8$ and $0.2$ for binary TTC and binary optical flow tasks respectively.
For training on the KITTI15 dataset, we use the same split as Yang and Ramanan~\cite{yang2020of2ttc}.

For the continuous version, we pre-train the network as before, and
fine-tune it on FlyingThings3D for continuous optical flow.
For each training image pair we uniformly sample horizontal and vertical shifts and stack the maps corresponding to each shift.
We use the resulting volumes to compute continuous optical flow via the AUC operation described in Section~\ref{sec:ttc_continuous}.
We can then continue training the network using a BCE loss on the individual probability maps, and a Smooth-L1 (SL1)~\cite{girshick2015fastrcnn} regression loss on the output of the AUC module.
We use relative weights of $0.1$ and $0.9$ respectively.
To fine-tune our network for continuous TTC estimation, we follow a similar strategy as for the binary segmentation training.
We continue training the network for the task of continuous optical flow and continuous TTC estimation on the Driving and the KITTI15 datasets.
However, this time we form three volumes, two corresponding to optical flow and one to TTC.
As discussed in Section~\ref{sec:inverse_ttc_domain}, we work in the inverse TTC domain and uniformly sample scale factors.
The network trained on the entire KITTI15 dataset is used for scene flow estimation on KITTI15 benchmark images, as explained in Section~\ref{sec:results}.
We refer the reader to our Supplementary for the additional training details.

\section{Evaluation and Results}\label{sec:results}

In this section we discuss qualitative and quantitative results for our method.
First, to validate the importance of the training strategy described in Section~\ref{sec:training_with_of}, we compare three training strategies: (1) we train our network only to estimate binary TTC, (2) we pre-train it to estimate binary optical flow (OF) and then binary TTC, and (3) we pre-train with binary OF, and then continue training with both binary OF and binary TTC (our method).
As shown in Table~\ref{table:binary_numerical}, our final method achieves a percentage error of $1.013$ and a mean intersection-over-union (mIOU) of $0.9525$.
If we only train for TTC estimation after pre-training for OF, we observe an increase in percentage error to $1.174$ and a drop in mIOU to $0.9453$.
Removing the binary OF estimation altogether, leads to an additional increase in percentage error to $2.670$ and an additional drop in mIOU to $0.8808$.

We also thoroughly validate our approach numerically on the KITTI15 validation set.
We compare to Yang and Ramanan~\cite{yang2020of2ttc}, who proposed the only existing approach that computes per-pixel TTC from a monocular camera, under practical assumptions.
They too look at how the size of objects changes, but they estimate it explicitly (and locally) from the optical flow between frames.
We also measure against PRSM~\cite{vogel20153d} and OSF~\cite{Menze2015kittidataset}, both of which perform 3D scene flow estimation using stereo cameras.
They represent images as a collection of planar super-pixels and jointly solve for geometry and 3D motion, which informs about the change of depth over time (Section~\ref{sec:data_ttc_from_sceneflow}).
This can be used to estimate motion-in-depth for each pixel, $\eta(x,y)$, using Equation~\ref{eq:ratio_of_depths}, which can then be thresholded.
Our final comparison is against Hur and Roth~\cite{hur2020selfmonosf}, a state-of-the-art monocular scene flow estimation approach.
We compare against their best model, which uses a combination of self-supervised learning on the KITTI15 raw dataset and supervised learning on the entire KITTI15 training dataset.
Note that this approach is already fine-tuned on our validation set.

Table~\ref{table:binary_numerical} shows that our continuous TTC estimation, yields the lowest motion-in-depth error:
\begin{equation}
    \text{MiD}=||\text{log}(\eta)-\text{log}(\eta_{GT})||_1 \cdot 10^4.
\end{equation}
However, our fundamental innovation is the ability to define a temporal geofence, \ie, detecting pixels with a TTC smaller than a given $\tau_i$ \emph{without the need to estimate the full TTC first}.
On this task we perform on par with Yang and Ramanan, but our binary TTC is around $26\times$ faster.
OSF performs better than our approach in terms of binary TTC, though, again, it uses a richer input.
Moreover, OSF and PRSM, implemented on a CPU, take $390$ s and $300$ s respectively.
The approach by Hur and Roth struggles despite fine-tuning the model on the validation set.
However, unlike ours and Yang and Ramanan's approaches, they do not use a synthetic dataset to pre-train, nor train for a better-posed TTC estimation task. Instead, they focus on the more difficult task of monocular scene flow estimation.

We further evaluate our method on the KITTI15 benchmark~\cite{menze2018JPRS} with the same strategy as Yang and Ramanan~\cite{yang2020of2ttc}:
we use our motion-in-depth estimation to compute scene flow, the 3D motion $[X(t_0), Y(t_0), Z(t_0)] \rightarrow [X(t_1), Y(t_1), Z(t_1)]$ corresponding to each pixel.
The first two scene flow components can be estimated by back-projecting the optical flow for each pixel.
For the $Z$ component, traditional approaches use stereo information.
We can estimate the third component from the motion-in-depth ratio if we are given the depth in one frame.
After computing the depth at $t_0$ with GANet~\cite{zhang2019ga} (evaluated by D1 in Table~\ref{table:kitti}) and combining it with our TTC estimate, we can compute the depth for each pixel at time $t_1$.
Therefore, D2 in Table~\ref{table:kitti} effectively measures the quality of our results.
The other numbers in the table evaluate other components and are reported for completeness.
While our method is not designed to estimate scene flow, it performs on par, or slightly better than methods specifically optimized for it.

In certain scenarios, binary TTC can be more informative than depth.
For instance, the top row of Figure~\ref{fig:teaser} shows a car driving away from the camera, and one that is farther, but driving towards the camera.
The latter, detected by our binary TTC, is potentially more impactful---in the true sense of the word!---for the ego vehicle path planning.
The last row of Figure~\ref{fig:binary_results} shows that, as discussed in Section~\ref{sec:ttc_binary}, our method can compensate for camera rotation, even if that breaks some of our assumptions.
This is also visible for continuous TTC in the first row of Figure~\ref{fig:optical_flow}.

We show quantized TTC estimation results in Figure~\ref{fig:teaser}\labelQ.
Note that even just $9$ TTC quantization levels ($8$ binary classifications) provide a meaningful representation of the scene.
Moreover, the underlying binary classifications can be run in parallel as they are independent of each other.
Therefore, quantized TTC can be run at roughly the same frame-rate as the binary TTC.

\begin{figure}
    \centering
    \includegraphics[width=\columnwidth]{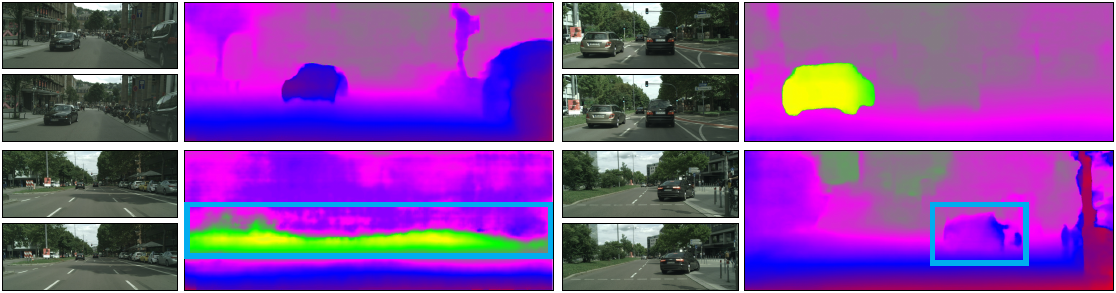}
    \vspace{-5mm}
    \caption{\textbf {Unseen dataset.} Our predicted TTC map on Citiscapes~\cite{cordts2016citiscapes}.
        The first row shows an example of a car moving towards us and another where the  car in the adjacent lane is speeding away.
        The bottom two rows show failures due to a road bump and a drastic rotation, respectively.}\label{fig:citiscapes_results}
    \vspace{-3mm}
\end{figure}

We show a visual comparison with Hur and Roth~\cite{hur2020selfmonosf}, and Yang and Ramanan~\cite{yang2020of2ttc} on continuous TTC for KITTI15 images in Figure~\ref{fig:all_results}.
Note the lower for our approach.
In Figure~\ref{fig:citiscapes_results} we show qualitative result of our approach on the Citiscapes dataset~\cite{cordts2016citiscapes}.
We show two failure cases on this dataset, one due to the sudden vertical motion caused by a road
bump and another due to an object rotating significantly.
Broadly, our method struggles for motions under-represented in the training dataset.

\begin{table}
    \setlength\tabcolsep{3.75pt}
    \centering
    \scriptsize{
        \begin{tabular}{cl|c|c|c}
                                    &                                       & \multicolumn{2}{c|}{Binary (200 ms -- 2 s)} & Continuous                                   \\
                                    &                                       & mIOU  ($\uparrow$)                          & \% error ($\downarrow$) & MiD ($\downarrow$) \\
            \hline
            \multirow{2}{*}{Stereo} & PRSM~\cite{vogel20153d}               & 0.9365                                      & 1.339                   & 124.0              \\
                                    & OSF~\cite{Menze2015kittidataset}      & \textbf{0.9556}                             & \textbf{0.941}          & 115.0              \\
            \hline
            \multirow{2}{*}{Mono}   & Hur \& Roth~\cite{hur2020selfmonosf}  & 0.9418                                      & 1.233                   & 115.13             \\
                                    & Yang \& Ramanan~\cite{yang2020of2ttc} & 0.9525                                      & 1.012                   & 75.00              \\
                                    & Ours                                  & 0.9525                                      & 1.013                   & \textbf{73.55}
        \end{tabular}
    }
    \vspace{-2mm}
    \caption{Comparison on the validation set of KITTI15 for both binary TTC (averaged over a set $\{\alpha_i\}$ uniformly sampled in the interval $\tau \in [0.02 \text{s}, 2 \text{s}]$) and continuous TTC. Note that PRSM and OSF both use richer input data (stereo vs mono). MiD, motion-in-depth, directly evaluates TTC. }
    \label{table:binary_numerical}
    \vspace{-3mm}
\end{table}

\begin{table}
    \setlength\tabcolsep{3.75pt}
    \centering
    \scriptsize{
        \begin{tabular}{l|c|c|c|c|c|c}
                                                & D1-all        & D2-bg         & D2-fg         & D2-all        & Fl-all        & SF-all        \\ \hline
            UberATG-DRISF~\cite{ma2019DRISF}    & 2.55          & 2.90          & 9.73          & 4.04          & 4.73          & 6.31          \\
            ACOSF~\cite{cong2020acosf}          & 3.58          & 3.82          & 12.74         & 5.31          & 5.79          & 7.90          \\
            ISF~\cite{behl2017sivapaper}        & 4.46          & 4.88          & 11.34         & 5.95          & 6.22          & 8.08          \\
            Yang\&Ramanan~\cite{yang2020of2ttc} & 1.81          & 3.39          & 8.54          & 4.25          & 6.30          & 8.12          \\
            \textbf{Ours}                       & \textbf{1.81} & \textbf{3.84} & \textbf{9.39} & \textbf{4.76} & \textbf{6.31} & \textbf{8.50}
        \end{tabular}
    }
    \vspace{-2mm}
    \caption{Top 5 published methods on the KITTI scene flow benchmark.
        Our method performs reasonably well, despite not being designed for scene flow, see text.}\label{table:kitti}
    \vspace{-0.05in}
\end{table}

\section{Conclusions}\label{sec:conclusions}

In certain scenarios, time-to-contact (TTC) information can be more useful than depth.
However, existing TTC estimation methods either make impractical assumptions, or cannot be run in real time.
We presented a framework to estimate time-to-contact (TTC) from a monocular input.
In just $6.4$ ms, our approach computes a temporal geofence to detect objects predicted to collide with the camera plane within a given TTC.
By computing a number of such geofences, it can also estimate TTC with arbitrary quantization, including continuous TTC.
We show that our method achieves competitive performance for TTC estimation---even when other methods use richer input data.

\vspace{-2mm}
\section*{Acknowledgments}
The authors would like to thank Stan Birchfield for the inspiring discussions on the theory of time-to-contact, Gengshan Yang and Junhwa Hur for their kind help in the comparisons with previous works, and anonymous reviewers and ACs for their helpful suggestions.

{\small
\bibliographystyle{ieee_fullname}
\bibliography{bittc}
}

\newpage
\clearpage
\twocolumn[
  \begin{@twocolumnfalse}
{
   \newpage
   \null
   \begin{center}
      {\Large \bf {B}inary {TTC}: {A} Temporal Geofence for Autonomous Navigation \\ ({S}upplementary)}
      {
      \large
      \lineskip .5em
      \begin{tabular}[t]{c}
          
      \end{tabular}
      \par
      }
      \vskip .5em
      \vspace*{0pt}
   \end{center}
}
  \end{@twocolumnfalse}
]


\setcounter{section}{0}
\setcounter{figure}{0}
\setcounter{table}{0}
\setcounter{footnote}{0}

\section{Additional Details on Architecture}\label{sec:arch}
\subsection{Binary Estimation}\label{sec:arch_bin}
Our binary segmentation architecture has three main components: a feature extraction network (FeatExtractNet), a segmentation network (SegNet2D), and a refinement network (SegRefineNet).
Our architecture blocks are adopted from Bi3DNet~\cite{badki2020Bi3D}.
We do not use any batch normalization layers in our network.

\paragraph{FeatExtractNet.}
We use the same feature extraction network as used in Bi3DNet~\cite{badki2020Bi3D}.
This feature extraction module is based on the simplified version of the feature extraction network from PSMNet~\cite{chang2018pyramid}.
We normalize each color channel of the input images using the mean and standard deviation of $0.5$ before passing to this network.
The output is a $32$-channel feature map at one third of the input image resolution.

\paragraph{SegNet2D.}
We warp the source image features using a task-dependent operator.
To avoid cropping out the image features during this warping operation, we zero pad the feature maps to $1.5\times$ the feature map resolution.
The reference image feature map and the warped source image feature map are concatenated and fed to the SegNet2D network.
SegNet2D is a 2D encoder-decoder network with skip-connections.
The encoder is comprised of five blocks, each of which has a conv layer that downsamples the features with a stride of 2 followed by another conv layer with a stride of 1.
We use $3\times3$ kernels.
The feature sizes for each of these five blocks are $128$, $256$, $512$, $1024$, and $1024$.
The decoder is comprised of five blocks, each of which has of a deconv layer with $4\times4$ kernels and a stride of 2, followed by a conv layer with $3\times3$ kernels and a stride of 1.
The feature sizes for each of these five blocks are $1024$, $512$, $256$, $128$, and $64$.
We use the LeakyReLU activation function in the network with a slope of $0.1$.
A final conv layer with $3\times3$ kernels, without any activation, is used to generate 3 outputs: one for binary TTC, one for binary horizontal optical flow, and one for binary vertical optical flow.
The final segmentation probability maps can be obtained by cropping out the excess padding, upsampling the outputs to the input image resolution and applying a sigmoid function.

\paragraph{SegRefineNet.}
SegRefineNet is used to refine the segmentation outputs from SegNet2D network using the reference image as a guide.
First, we generate a $16$ channel feature map for the reference image by applying 3 conv layers with $3\times3$ kernels and a stride of 1.
The first two layers use ReLU activation and the final layer does not use any activation.
This feature extraction is done only once and can be used for refining multiple segmentation outputs from the SegNet2D network.
An upsampled segmentation output is concatenated with the reference image features and refined by applying 4 conv layers.
Each conv layer uses $3\times3$ kernels, a feature-size of $8$, and LeakyReLU activation with a slope of 0.1.
The final output of this network is generated by a final conv layer with $3\times3$ kernel and without any activation.
The final binary segmentation probability map can then be generated by applying a sigmoid function.

\subsection{Continuous Estimation}\label{sec:arch_cont}
Given input images we generate the corresponding feature maps using the same FeatExtractNet as in Section~\ref{sec:arch_bin}.
To generate continuous estimation results we uniformly sample the warping parameters in the desired range and use these parameters to warp the features of the source image.
These warped source image features are concatenated with the reference image features to form an input volume for the SegNet2D network.
SegNet2D generates an output volume which upon upsampling to the input image resolution, applying the sigmoid operator followed by AUC operator gives us the continuous estimation map.
This continuous map is further refined using the input image as a guide using a refinement network, ContRefineNet.
This network is based on the disparity refinement network proposed in StereoNet~\cite{khamis2018stereonet}.

\section{Additional Details on Training}\label{sec:train}

\subsection{Binary Estimation}
We pre-train our network for the binary optical flow task on the FlyingChairs2~\cite{dosovitskiy2015flownet, iig2018generic} dataset for 300 epochs and train it further on FlyingThings3D dataset~\cite{mayer2016large} for 400 epochs using a binary cross-entropy (BCE) loss with respect to a thresholded version of the ground truth.
Each batch for training is formed by randomly sampling 16 image pairs from the dataset and then randomly sampling shifts for the two components of the optical flow vector for each image.
Note that our SegNet2D network has three output heads and we leave the head corresponding to binary TTC untouched during this training.
We then fine-tune our network for estimating binary TTC first on the Driving~\cite{mayer2016large} for 500 epochs and then on the KITTI15~\cite{Menze2015kittidataset} datasets for 10k epochs.
Since both datasets also offer optical flow data, in this second stage we train for both binary optical flow and binary TCC.
To do this we form a batch by randomly sampling 16 image pairs from the dataset.
Then for each image pair we form two sets of warped image features: one by randomly sampling shifts for the two components of the optical flow vector and other by randomly sampling scales for training binary TTC.
We select the appropriate segmentation output corresponding to the task and use BCE loss to train our network simultaneously for the binary optical flow and TTC estimation task.
Throughout our training we randomly sample shifts in the range $[-99,99]$, and scales in the range of $[0.5,1.3]$.
We use relative weights of $0.8$ and $0.2$ for binary TTC and binary optical flow task respectively.
For training on the KITTI15 dataset, we split our dataset into training ($160$ examples) and validation ($40$ examples) sets, following the split provided by Yang and Ramanan~\cite{yang2020of2ttc}.

\subsection{Continuous Estimation}

To train our network for the continuous estimation task, we start with the network trained on the FlyingChairs2 and FlyingThings3D datasets for the binary optical flow task and fine-tune it on FlyingThings3D for continuous optical flow for 100 epochs.
Each batch for training is formed by randomly sampling 8 image pairs from the dataset.
For each training image pair we uniformly sample horizontal and vertical shifts and stack the maps corresponding to each shift.
We use the resulting volumes to compute continuous optical flow via the AUC operation which we refine using ContRefineNet.
To fine-tune our network for continuous TTC estimation, we follow a similar strategy as for the binary segmentation training.
We continue training the network for the task of continuous optical flow and continuous TTC estimation on Driving dataset for 100 epochs, followed by KITTI15 datasets for 600 epochs.
We form three volumes, two corresponding for optical flow and one for TTC.
Throughout the training, we use BCE loss on the estimated binary segmentation probability maps, and a SmoothL1 (SL1) regression loss on the output of the AUC module.
We use relative weights of $0.1$ and $0.9$ respectively.
While training on both TTC and OF estimation tasks, we use relative weights of $0.8$ and $0.2$ respectively.
For optical flow we randomly sample a contiguous block of 16 shifts that are divisible by 3 and in the range $[-99, 99]$ for both the components of the optical flow.
We select the shifts to be a factor of 3 since we perform the shifting on the feature maps that are one third the input image resolution.
The AUC operation seamlessly handles the objects with shifts that lie beyond sampled range.
For TTC estimation, we work in the inverse TTC domain and uniformly sample 24 scales in the range $[0.5, 1.3]$.
We perform the training on the cropped images of size $384\times576$.
The cropping is done after the feature warping step.
Again for the KITTI15 dataset, we split the dataset into train and validation sets and use the validation set to compare our method with competing approaches.
The network trained on the entire KITTI15 dataset is used for scene flow estimation on KITTI15 benchmark images as explained in the main paper.

\end{document}